\documentclass[conference]{IEEEtran}
\IEEEoverridecommandlockouts
\usepackage{cite}
\usepackage{amsmath,amssymb,amsfonts}
\usepackage{algorithmic}
\usepackage{graphicx}
\usepackage{textcomp}
\usepackage{xcolor}
\def\BibTeX{{\rm B\kern-.05em{\sc i\kern-.025em b}\kern-.08em
    T\kern-.1667em\lower.7ex\hbox{E}\kern-.125emX}}

\usepackage{graphicx}
\usepackage{amsmath}
\usepackage{amssymb}
\usepackage{booktabs}
\usepackage[T1]{fontenc}
\usepackage[pagebackref,breaklinks,colorlinks]{hyperref}

\usepackage[capitalize]{cleveref}
\crefname{section}{Sec.}{Secs.}
\Crefname{section}{Section}{Sections}

\usepackage{newtxtext}
\usepackage[varg]{newtxmath}
\usepackage{bm}
\usepackage{multirow}
\usepackage{subcaption}
\usepackage{comment}
\usepackage{color}
\usepackage{arydshln}

\DeclareMathOperator{\act}{{\bf a}}
\DeclareMathOperator{\z}{{\bf z}}
\DeclareMathOperator{\q}{{\bf q}}
\DeclareMathOperator{\hvec}{{\bf h}}
\DeclareMathOperator{\s}{{\bf s}}
\DeclareMathOperator{\e}{{\bf e}}
\DeclareMathOperator{\dir}{{\bf d}}
\DeclareMathOperator{\mub}{{\bf \mu}}
\newcommand{\first}[1]{\bf{{\underline{#1}}}}
\newcommand{\second}[1]{\bf{#1}}

\begin{document}

\title{
Future Predictive Success-or-Failure Classification \\for Long-Horizon Robotic Tasks
}

\author{
\IEEEauthorblockN{Naoya Sogi, Hiroyuki Oyama, Takashi Shibata, and Makoto Terao}
\IEEEauthorblockA{
NEC Corporation, \\
Kanagawa, Japan\\
Email: \{naoya-sogi, h.oyama\}@nec.com, t.shibata@ieee.org, m-terao@nec.com}
}

\maketitle

\begin{abstract}
Automating long-horizon tasks with a robotic arm has been a central research topic in robotics. Optimization-based action planning is an efficient approach for creating an action plan to complete a given task. Construction of a reliable planning method requires a design process of conditions, e.g., to avoid collision between objects. The design process, however, has two critical issues: 1) iterative trials--the design process is time-consuming due to the trial-and-error process of modifying conditions, and 2) manual redesign--it is difficult to cover all the necessary conditions manually. To tackle these issues, this paper proposes a future-predictive success-or-failure-classification method to obtain conditions automatically. The key idea behind the proposed method is an end-to-end approach for determining whether the action plan can complete a given task instead of manually redesigning the conditions. The proposed method uses a long-horizon future-prediction method to enable success-or-failure classification without the execution of an action plan. This paper also proposes a regularization term called transition consistency regularization to provide easy-to-predict feature distribution. The regularization term improves future prediction and classification performance. The effectiveness of our method is demonstrated through classification and robotic-manipulation experiments.
\end{abstract}

\section{Introduction}
\label{sec:intro}
Long-horizon tasks are a set of complex tasks that require the execution of multiple actions to complete the task such as objects replacement and stacking. Automating such complex tasks with a robotic arm has been a central research topic in robotics, as various real-world applications are categorized as long-horizon tasks. Therefore, various action-planning methods have been proposed~\cite{Kaelbling2011HierarchicalTA,janner2022diffuser,shahValueFunctionSpaces2022,driess2020deep} for creating an action plan consisting of a sequence of action categories and input signals to the robotic arm while taking into account long-horizon changes in the environment.

\begin{figure}[tb]
    \centering
      \begin{minipage}[b]{0.97\linewidth}
        \centering
        \includegraphics[width=0.97\linewidth]{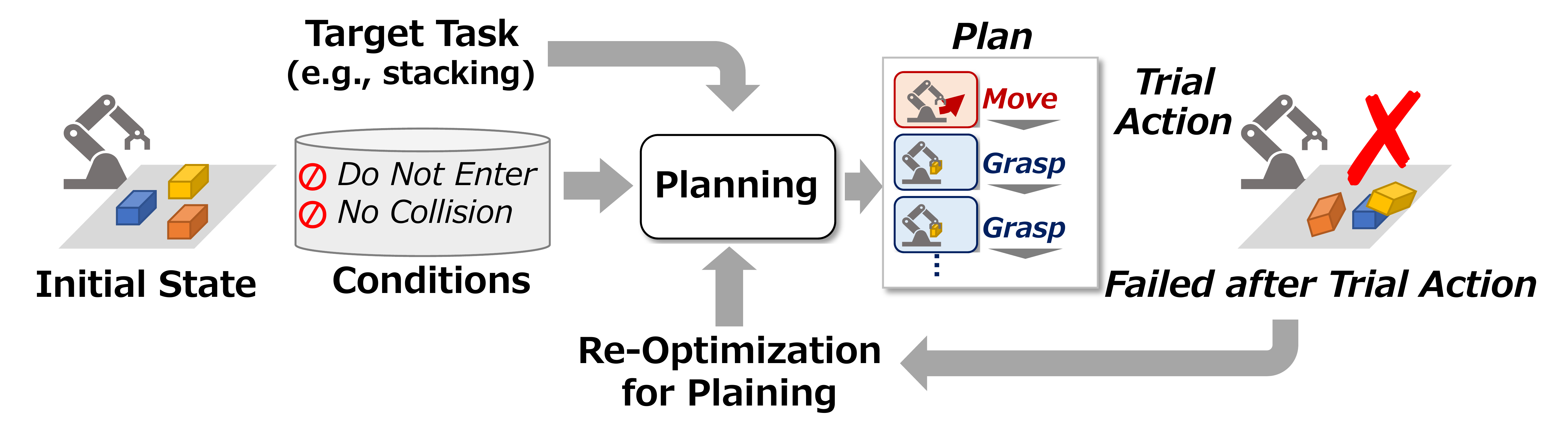}
        \subcaption{Optimization-based planning methods.}
        \label{fig:abst_a}
    \end{minipage} \\ \vspace{2mm}
    \begin{minipage}[b]{0.97\linewidth}
        \centering
        \includegraphics[width=0.97\linewidth]{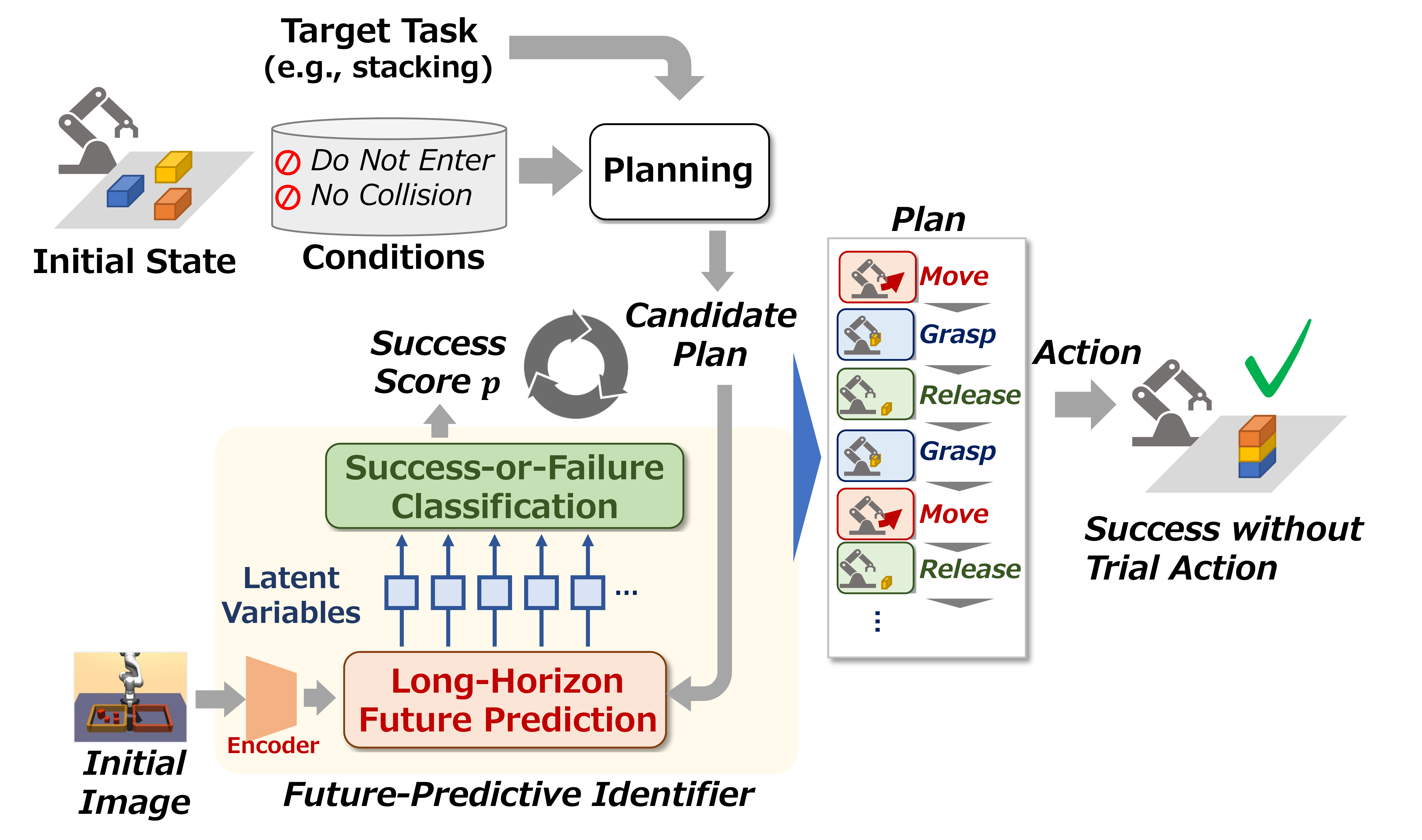}
        \subcaption{Integration of our method and a planning method.}
        \label{fig:abst_b}
    \end{minipage}
    \caption{Conceptual diagram of the proposed method. (a) Performance of optimization-based planning methods depends on manually designed conditions. (b) The proposed success-or-failure classification method supports or replace the conditions.}
    \label{fig:abst}
\end{figure}

Task and motion planning (TAMP) is an efficient planning framework for long-horizon tasks~\cite{Kaelbling2011HierarchicalTA,Hartmann2020RobustTA, driess2020deep,takano2021continuous}. TAMP can be used to create a highly feasible action plan by first creating a sequence of action categories to take into account long-horizon changes. Representative TAMP-based methods are optimization-based methods that formulate the planning problem as an optimization problem and create an action plan by solving the optimization problem. Optimization-based methods require conditions, as shown in~\cref{fig:abst_a}, to avoid falling into unreasonable states each time, such as collisions between objects, in addition to the initial position of objects and the task objective.

Although the design of conditions is essential for the construction of reliable planning methods, the design of conditions has two critical issues: 1) iterative trials--the design process is time-consuming, as it requires a trial-and-error process of modifying conditions while checking the naive operation of the robotic arm, and 2) manual redesign--it is difficult to manually cover all the necessary conditions since the number of situations to be considered increases as tasks become more complex. Thus, reliable planning methods are required without the iterative trials and manual redesign.

\footnote[0]{©2024 IEEE. Personal use of this material is permitted. Permission from IEEE must be obtained for all other uses, in any current or future media, including reprinting/republishing this material for advertising or promotional purposes, creating new collective works, for resale or redistribution to servers or lists, or reuse of any copyrighted component of this work in other works.}

To tackle these issues, this paper proposes a future-predictive success-or-failure-classification method called {\bf{F}}uture-predictive {\bf{I}}dentifier for {\bf{R}}obot {\bf{P}}lanning (FIRP) to obtain conditions from data automatically. It is an end-to-end method that determines whether the action plan can complete a given task instead of manually redesigning the conditions. FIRP involves the following two steps: 1) executing a long-horizon transition prediction of image features obtained by implementing an action plan then 2) outputting success-or-failure scores with the predicted image features. This two-step method enables us to carry out success-or-failure classification without implementing an action plan. As shown in \cref{fig:abst_b}, iterative trials and manual redesign become unnecessary by executing re-planning when FIRP identifies an action plan that is not feasible. To the best of our knowledge, this is the first study to propose a future-predictive success-or-failure-classification method for long-horizon tasks in contact-rich situations with a robot and the environment.

Robust long-horizon prediction is essential for FIRP. For such a challenging task, we use the recurrent state space model (RSSM)~\cite{planet}, a prediction model that takes into account the uncertainty of transitions by introducing probabilistic features. This paper also proposes a regularization term called transition consistency regularization (TCR). TCR is based on two consistencies: 1) temporal transition consistency (TTC) and 2) action-transition consistency (ATC). These consistencies are inspired by the simple idea that long-horizon transitions can be decomposed into elements common to the temporal direction and action categories. TCR provides a more predictable feature distribution, enabling accurate prediction and classification.

Our main contributions are summarized as follows:
\begin{enumerate}
\item This paper proposes a future-predictive success-or-failure-classification method to automatically obtain conditions required by planning methods.
\item This paper proposes a regularization term to improve long-horizon prediction and classification accuracy.
\item The effectiveness of our method is demonstrated through classification and robotic-manipulation experiments.
\end{enumerate}

\begin{figure*}[thb]
    \centering
    \includegraphics[width=0.77\linewidth]{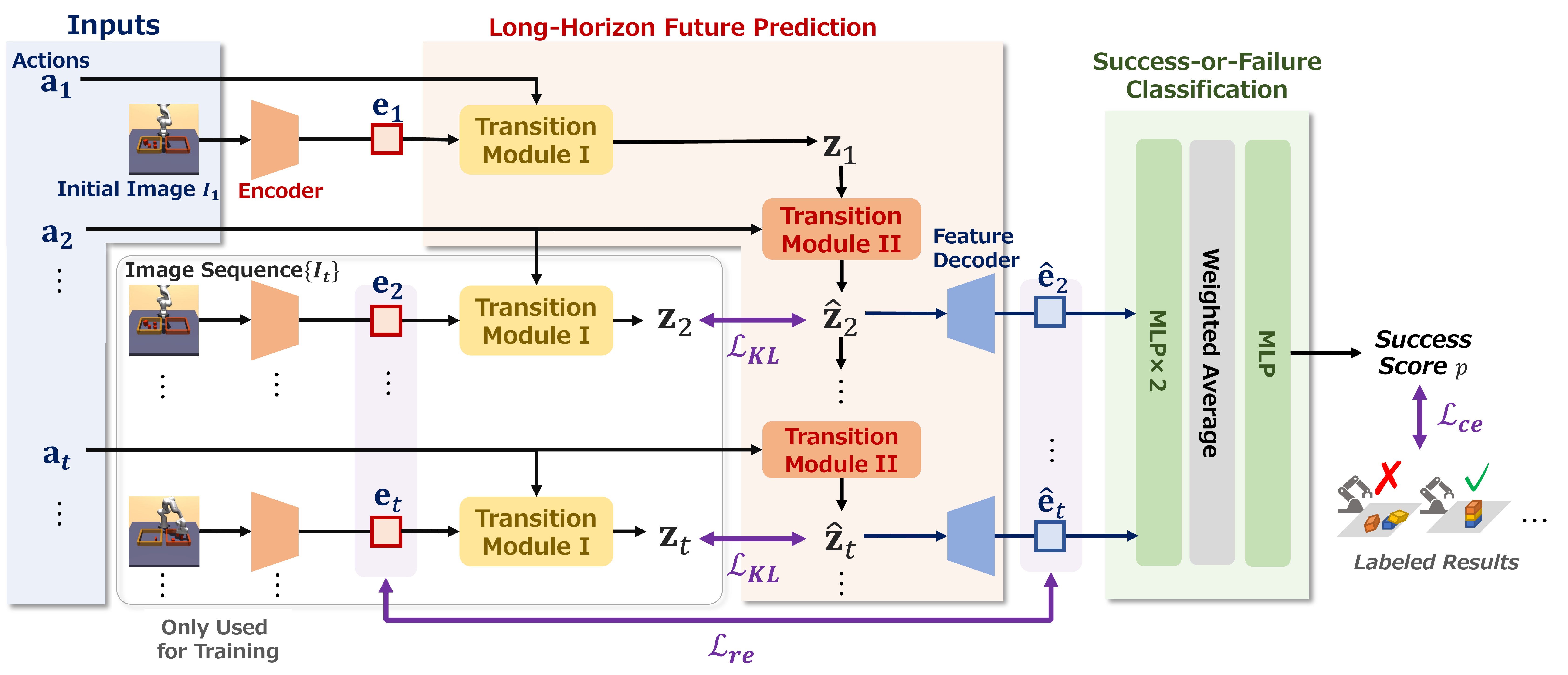}
    \caption{Processing pipeline of the proposed method called Future-predictive identifier for robot planning; FIRP. FIRP receives an initial image $I_1$ and actions $\{\act_t\} $, then outputs a success score $p$ using long-horizon prediction of subsequent image features $\{\e_t\}_{t=2}^T$.
    The image feature prediction is based on latent variables $\{\z_t\}$ behind image features.
    FIRP is trained using image sequences and three loss functions ${\cal L}_{KL}$, ${\cal L}_{re}$ and ${\cal L}_{ce}$: the first two are for the prediction, and the last is for the classification.}
    \label{fig:proposed}
\end{figure*}

\begin{figure}[thb]
    \centering
    \includegraphics[width=0.75\linewidth]{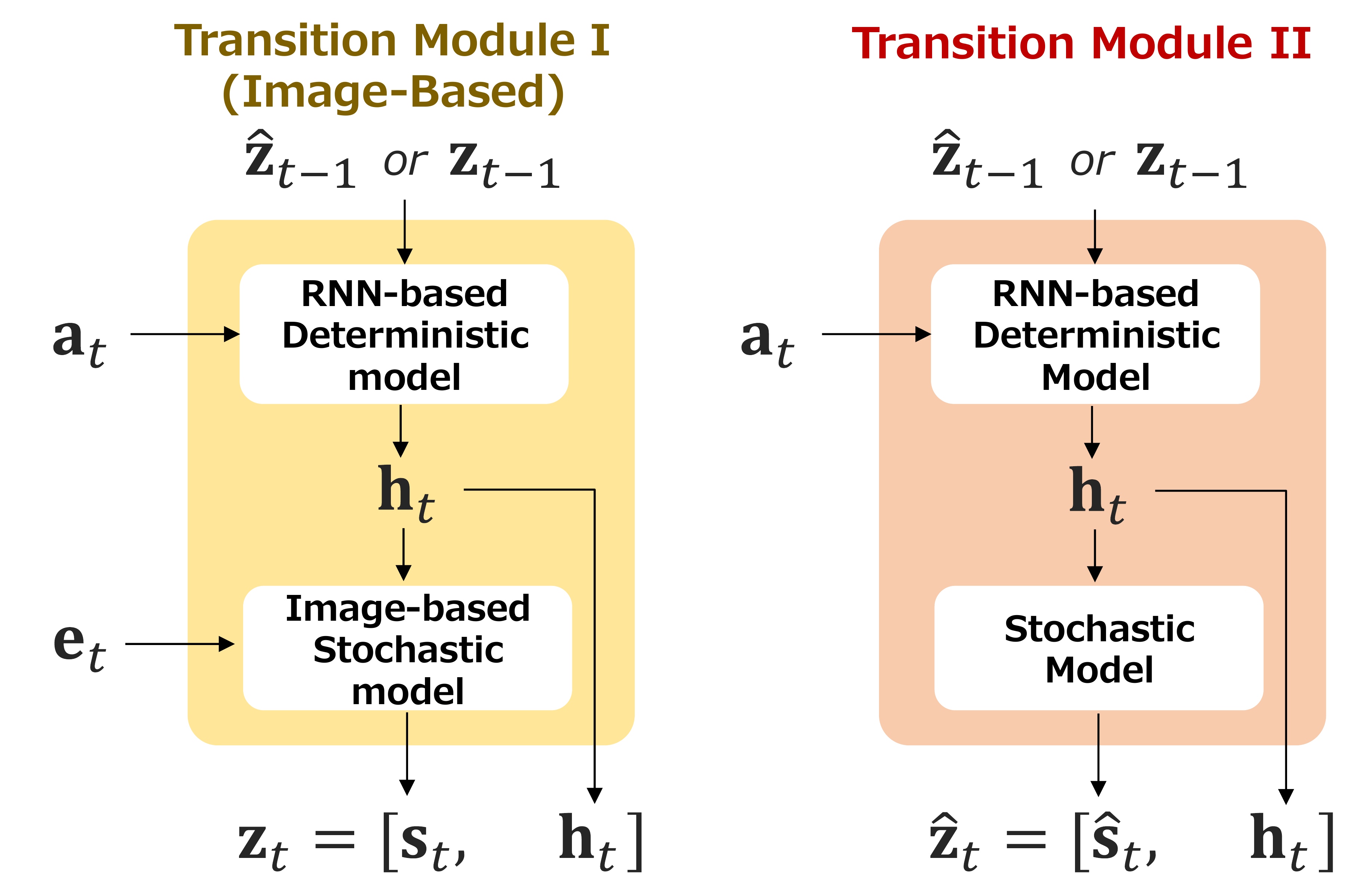}
    \caption{Pipeline of the future prediction of latent variables $\{\z_t\}$.
    A latent variable $\z_t$ consists of a stochastic and deterministic feature and is used to decode an image feature.}
    \label{fig:proposed_rssm}
\end{figure}

\section{Related Work}
\label{sec:related}
This section overviews applications of success-or-failure classification in robotics, and a future prediction method, RSSM.

\subsection{Failure-or-Success Classification}
\label{sec:related_fs}
We briefly review three types of applications of success-or-failure classification and highlight our contributions.

\noindent{\bf i.~Failure recovery:~}
Failure recovery is a task to recover from an error during an automatic robotics process. A failure-classification method determines whether a recovery process should be started~\cite{FINONetDeepMultimodal,LearningRelationalAffordance,WhatWentWrong}. Such methods assign a failure score while sensing the current state during the robotic-automation process. The main streams are supervised classification methods and anomaly-detection-based methods.

\noindent{\bf ii.~Parameter estimation:~}
Parameter estimation is a process to estimate the parameters required to execute short-term tasks~\cite{RoboticGraspDetection,zeng2022robotic,lenz2015deep,6DOFGraspNetVariational}. For instance, a three or six degrees-of-freedom grasping pose is estimated using a grasp success evaluator. There are two types of evaluators: 1) an evaluator that receives a pose candidate then outputs the success score of the candidate~\cite{6DOFGraspNetVariational}, and 2) an evaluator that receives an image reflecting an environment then outputs a grasp-feasibility map; each element is a success score when a grasp is executed at the corresponding pixel location~\cite{zeng2022robotic}.

\noindent{\bf iii.~Action planning:~}
A success-or-failure classification method is also used for action-planning methods~\cite{levine2018learning,SkillLearningTask,LearningbasedSuccessValidation,chen2021learning,MotionPlanningSuccess,VisuallyGrounded}. Various methods have been proposed for short-term tasks, such as object grasping, pushing, and moving. These methods differ in the input for a classifier, such as an action sequence~\cite{MotionPlanningSuccess}, action and image~\cite{levine2018learning}, or action sequence and images~\cite{chen2021learning}.
GROP~\cite{VisuallyGrounded} is one of the few examples using a success-or-failure classifier for a long-horizon task.

\noindent {\bf Our contributions:~}
In the above three applications, our method is categorized as action planning.
The most related concept is GROP~\cite{VisuallyGrounded}, which uses an action feasibility map for a long-horizon task.
However, GROP can not be applied to our setting, as we tackle a more challenging setting in two aspects: we assume that 1) a robotic arm interacts richly with the environment, and 2) our method needs to predict the future environment induced by robotic actions, unlike the GROP's setting, which can access the actual environment.

\subsection{Recurrent State Space Model}
\label{sec:rssm}
Recurrent state space model (RSSM) is a future prediction model that can perform accurate long-horizon predictions~\cite{planet}. 
The effectiveness is due to the introduction of probabilistic features to consider stochastic transition occurring in the real world.
RSSM is evaluated in relatively simple tasks with the sampling-based planning method~\cite{planet,cem}.
Subsequently, the extension of RSSM, e.g., novel computation of probabilistic features and learning method, has been proposed and combined with model-based reinforcement learning to extend its range of applications~\cite{dreamer,dreamerv2}.
This paper proposes a novel regularization term to exploit the capability of RSSM fully.

\section{Proposed Method}
\label{sec:proposed}
Our FIRP framework aims to determine whether the action plan can complete a given task instead of manually redesigning the conditions.
FIRP consists of two processes: 1) future prediction and 2) success-or-failure classification, as shown in~\cref{fig:proposed}.
For future prediction, FIRP uses RSSM~\cite{planet}, which enables accurate long-horizon prediction.
Then, success-or-failure classification is executed with the predicted features.
Our proposed time consistency regularization (TCR) provides a more predictable feature distribution, enabling stable and accurate prediction and classification.
We first describe the problem setting and then explain FIRP and its basic learning algorithm. Finally, we introduce our regularization term, TCR.

\label{sec:proposed_firp}

\subsection{Future-Predictive Success-or-Failure Classification}
\subsubsection{Problem Settings}
Let $I_1$ be an RGB image of the initial environment (hereafter, we refer to $I_1$ as an initial image) and $\{\act_t\}_{t=1}^{T}$ be an action plan to execute a long-horizon task. The $t$-th action $\act_t$ consists of input signals to robotic joint angles and an executing action category, such as grasp, move, and release.
FIRP predicts whether the given task will succeed, using the initial image $I_1$ and the action plan $\{\act_t\}_{t=1}^{T}$ without executing the action plan in an actual environment.

Our framework first extracts an image feature $\e_1$ from $I_1$ then predicts subsequent image features $\{{\hat \e}_t\}_{t=2}^T $ by using the action plan $\{\act_t\}_{t=1}^{T}$.
Finally, the predicted feature sequence is used to to identify success or failure.
Note that, for clarity of notation, predicted values are distinguished by hat, as in ${\hat \e}_t$.

\subsubsection{Long-Horizon Future Prediction}
We use RSSM~\cite{planet} to carry out the prediction of $\{{\e}_t\}_{t=2}^T$.
RSSM first executes the transition of latent variables $\{\z_t\}$ behind $\{{\e}_t\}$, then infers image features $\{{\e}_t\}_{t=1}^T$ from the latent variables.
A latent variable $\z_t$ consists of a deterministic feature $\hvec_t$ and stochastic feature $\s_t$.
As shown in ~\cref{fig:proposed_rssm}, the prediction process of RSSM is given as:
\begin{align}
    \label{eq:rssm}
    {\rm Deterministic~model:~} \hvec_t &= {\rm RNN}(\hvec_{t-1}, f(\s_{t-1}, \act_t)), \\
    \label{eq:rssm2}
    {\rm Transition~module~I~:~} \s_t &= g(\hvec_t, \e_t), \\
    \label{eq:rssm3}
    {\rm Transition~module~I\hspace{-.01em}I~:~} {\hat \s}_t &= {\hat g}(\hvec_t), \\
    {\rm Feature~decoder:~}{\hat\e}_t&= 
    \begin{cases}
        {{D}(\s_t, \hvec_t) ~I_t~is~available,} \\
        {{D}({\hat \s}_t, \hvec_t)~otherwise,}
    \end{cases}
    \label{eq:rssm4}
\end{align}
where ${\rm RNN}$ is a recurrent neural network, $f, g, {\hat g}, D$ are multi-layer perceptrons (MLPs) with two fully connected layers and rectified-linear-unit activation, and $g, {\hat g}$ outputs a stochastic feature by the reparametrization trick~\cite{vae}. Gaussian distribution is used for this trick.
This stochastic feature is a key of RSSM to perform accurate long-horizon predictions.
Transition module I is used to transition of deterministic latent variables, when an image $I_t$ is available. 
Transition module $\rm I\hspace{-.01em}I$ is used to predict future latent variables. 
We set initial values $\s_0, \hvec_0$ to a zero vector, respectively, as in a previous study~\cite{planet}.

\subsubsection{Success-or-Failure Classification}
For success-or-failure classification without the execution of an action plan, we first obtain predicted image features $\{{\hat \e}_t\}_{t=2}^T$ by giving the initial image $I_1$ and $\{\act_t\}_{t=1}^T$ to \cref{eq:rssm,eq:rssm2,eq:rssm3,eq:rssm4}.
We then calculate a success score after unifying image features $\e_1, \{{\hat \e}_t\}_{t=2}^T$

To unify image features, we first transform the image features using a two-layer MLP, i.e., $\q_t={\rm MLP}({\hat \e}_t)$.
We then unify the transformed features by the weighted sum.
The weights $\{w_t\}$ are calculated using the softmax function with temperature and the cosine similarity between ${\q}_t$ and learnable parameters $\{\phi_i\}_{i=1}^M$ as follows:
\begin{align}
    w_t = \frac{1}{M}\sum_{i=1}^M\frac{e^{\cos(\phi_i, {\q}_t) / L}}{\sum_{j=1}^T{e^{\cos(\phi_i, {\q}_j) / L}}}, 
    \label{eq:weights}
\end{align}
where $M$ is the number of learnable parameters, and $L$ is the temperature parameter, which in this study was set to 20.

The feature obtained by the weighted sum is used as input of a fully connected layer which calculates the success score $p$.

\subsubsection{Learning Algorithm}
\label{sec:proposed_learning}
Suppose that there are $N$ training samples, i.e., $N$ action plans $\{\act_t^n\}_{n=1}^N$ and corresponding image sequences $\{I_t^n\}$ obtained by executing the action plans. We also assume that a task success or failure label is given to each training sample. Two types of loss functions are used to train FIRP: 1) future-prediction loss and 2) classification loss.

\noindent {\bf 1.~Future prediction loss ${\cal L}_{f}^\tau$ :}
Future prediction loss ${\cal L}_{f}^\tau$ consists of two terms, ${\cal L}_{KL}^\tau$ and ${\cal L}_{re}^\tau$, as follows:
\begin{align}
    {\cal L}_{f}^\tau={\cal L}_{KL}^\tau+{\cal L}_{re}^\tau,
\end{align}
where $\tau$ is the number of prediction steps.

The first loss term ${\cal L}_{KL}^\tau$ is to accurately predict stochastic latent variables $\{\s_t^n\}$.
As $\s_t^n$ is generated by the reparametrization trick using a Gaussian distribution, we use the Kullback-Leibler (KL) divergence loss to evaluate prediction error:
\begin{align}
    {\cal L}_{KL}^\tau = \frac{1}{NT}\sum_{t,n}{\rm KL}(\s_t^n\| {\hat \s}_t^n),
\end{align}
where, $\{\s_t^n\}$ is calculated using image sequences and \cref{eq:rssm2}, and $\{\hat \s_t^n\}$ is predicted from $\hvec_{t-\tau}$ using \cref{eq:rssm,eq:rssm3}.

The second loss term ${\cal L}_{re}^\tau$ is to match a target image feature $\e_t^n$ and decoded feature ${\hat \e}_t^n$ from ${\hat \s}_t^n$.
We simply use the mean square error ${\cal L}_{re}^\tau$ between a target image feature and decoded feature as follows:
\begin{align}
    {\cal L}_{re}^\tau = \frac{1}{NT}\sum_{n=1}^N{\sum_{t=2}^T{\|\e_t^n - {\hat \e}_t^n \|_2}}.
    \end{align}

\noindent {\bf 2.~Classification loss ${\cal L}_{ce}$ :}
Our FIRP is also trained using classification loss ${\cal L}_{ce}$ so that an output success score $p$ matches the success/failure label.
In the training procedure, ${\cal L}_{ce}$ is calculated from success scores estimated from image features $\{\e_t^n\}$.
Both ${\cal L}_{f}^1$ and ${\cal L}_{ce}$ are used in the basic learning algorithm.

\noindent {\bf3.~Latent overshooting:} FIRP uses an additional loss term, latent overshooting~\cite{overshooting,planet}, for accurate long-horizon prediction. In latent overshooting, a stochastic feature ${\hat\s}_t^n$ is predicted using \cref{eq:rssm,eq:rssm3} from latent variables, $\hvec_{t-\tau}^n, \s_{t-\tau}^n $, at $\tau(>1)$ time before and used to calculate ${\cal L}_f^\tau$.

\noindent {\bf Loss function:} The overall loss function is as follows:
\begin{equation}
    {\cal L}_{ce} + {\cal L}_f^1 +\frac{\lambda}{T-1}\sum_{\tau=2}^{T-1}{({\cal L}_{KL}^\tau + {\cal L}_{ce}^\tau )},
    \label{eq:loss}
\end{equation}
where $\lambda$ is a positive value to tune the effect of latent overshooting. FIRP is learned on the basis of this loss function.

\subsection{TCR:~Transition Consistency Regularization}

\label{sec:proposed_reg}
\begin{figure}[tb]
    \centering
     \begin{minipage}[b]{0.97\linewidth}
        \centering
        \includegraphics[width=0.6\linewidth]{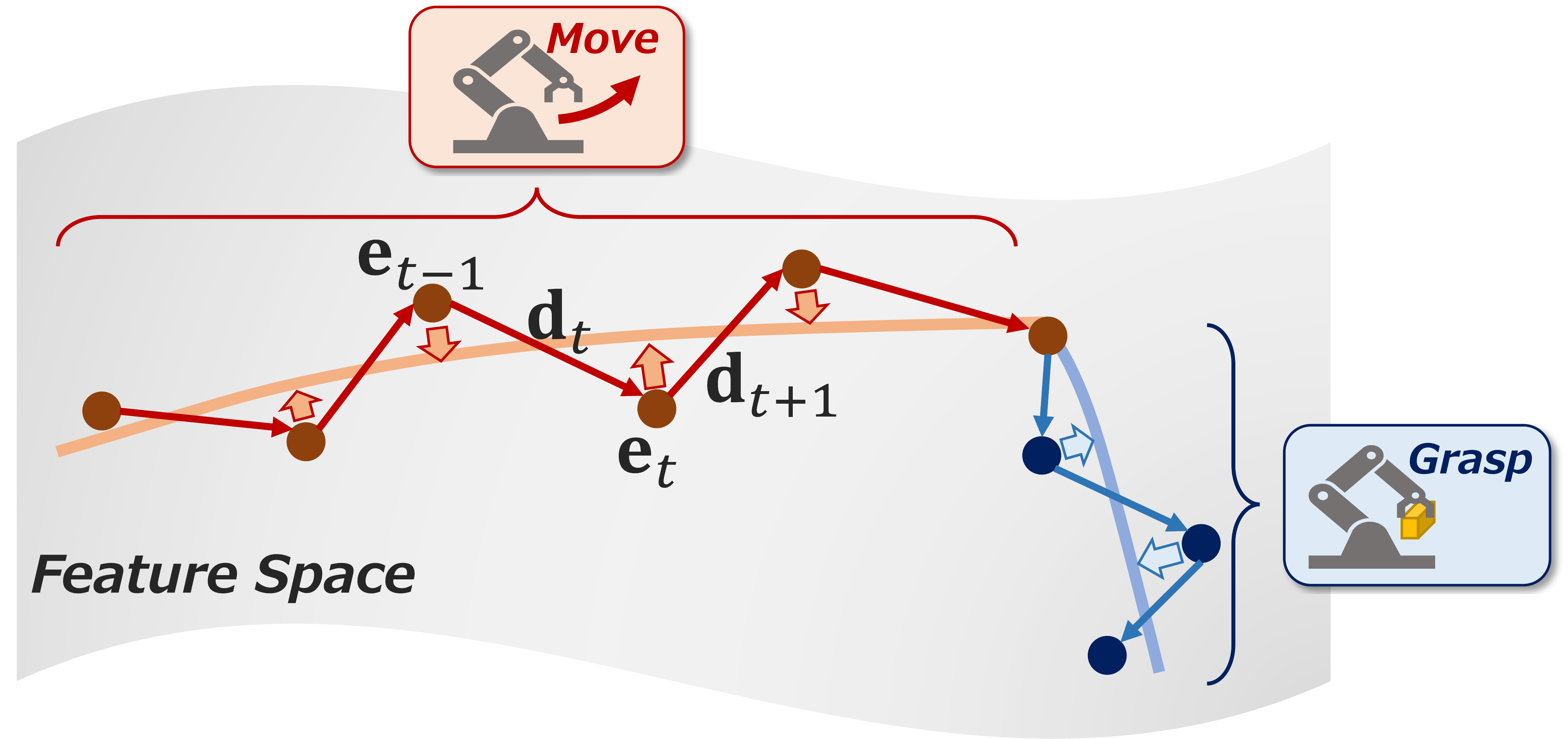}
        \subcaption{TTC maintains consistency in temporal transition by suppressing the explosive temporal change of image features.
        }
        \label{fig:ttc}
    \end{minipage}
     \begin{minipage}[b]{0.97\linewidth}
        \centering
        \includegraphics[width=0.6\linewidth]{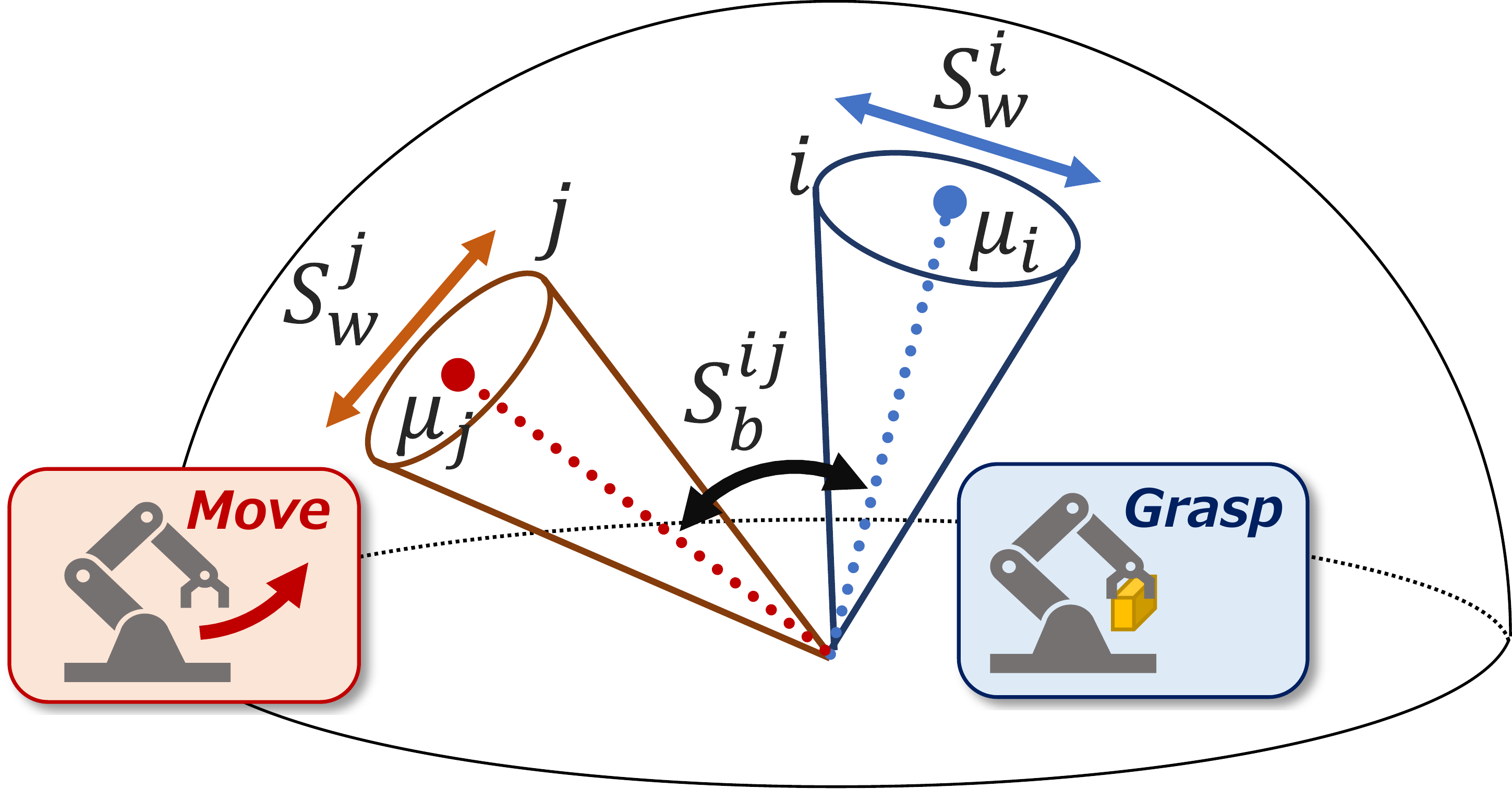}
        \subcaption{ATC maintains consistency between actions and transition directions, i.e., making transition directions in the same action category similar while making transition directions in the different categories dis-similar.}
        \label{fig:atc}
    \end{minipage}
    \caption{Conceptual diagram of transition consistency regularization (TCR). TCR consists of (a) temporal transition consistency (TTC) and (b) action-transition consistency (ATC). }
    \label{fig:proposed_reg}
\end{figure}

Robust long-horizon prediction is essential for FIRP. For such a challenging task, our proposed regularization term TCR is used to produce a more predictable transition of image features $\{\e_t\}$, enabling stable and accurate prediction and success-or-failure classification. 
As shown in \cref{fig:proposed_reg}, TCR consists of two consistencies: (a) \textbf{T}emporal \textbf{T}ransition \textbf{C}onsistency \textbf{(TTC)} and (b) \textbf{A}ction-\textbf{T}ransition \textbf{C}onsistency (\textbf{ATC}). 
Formally, TCR is defined as follows:
\begin{align}
    R_{TCR} = \alpha R_{TTC} + \beta R_{ATC}, 
    \label{eq:tcr}
\end{align}
where $R_{TTC}$ and $R_{ATC}$ are regularization terms based on TTC and ATC, respectively.
$\alpha$ and $\beta$ are positive values to balance the effect of TTC and ATC.

In the following discussion on TTC and ATC, we use a term of a transition direction, $\dir_t=\e_t-\e_{t-1}$, which is induced by an action $\act_t$. 

\subsubsection{TTC:~Temporal Transition Consistency}
Temporal transition consistency (TTC) maintains consistency in temporal transition by suppressing the explosive temporal change of image features, as shown in \cref{fig:ttc}. This means that transition directions $\{\dir_t\}_t $ do not change abruptly or are always the same. TTC could improve future prediction performance, as shown in our experiment (\cref{sec:cls_exp}).

TTC increases smoothness and sparseness in a temporal change of transition directions. Smoothness is to suppress abrupt change, and sparseness is to suppress occurring changes of transition directions, respectively.
We use the following equations: 
\begin{align}
    {\rm Smoothness}:~ R_{sm}&=\sum_{t=3}^{T-1} \|2\theta_t-\theta_{t-1}-\theta_{t+1} \|_2^2=\sum_{t=3}^{T-1}\|c_t\|_2^2,     \label{eq:sm} \\
    {\rm Sparseness}: ~ R_{sp}&=\sum_{t=3}^{T-1} |c_t|,
    \label{eq:sp}
\end{align}
where $\theta_t$ is an angle between the transition direction $\dir_t$ and $\dir_{t-1}$, and $c_t(=2\theta_t-\theta_{t-1}-\theta_{t+1})$ is used for evaluating temporal-change degrees.
TTC is the sum of \cref{eq:sm,eq:sp} as follows:
\begin{align}
    {\rm TTC}: ~R_{TTC}= R_{sm} + R_{sp}.
    \label{eq:ttc}
\end{align}
By minimizing \cref{eq:ttc}, consistency in temporal transition is maintained. 

\begin{figure}
    \centering
    \includegraphics[width=0.58\linewidth]{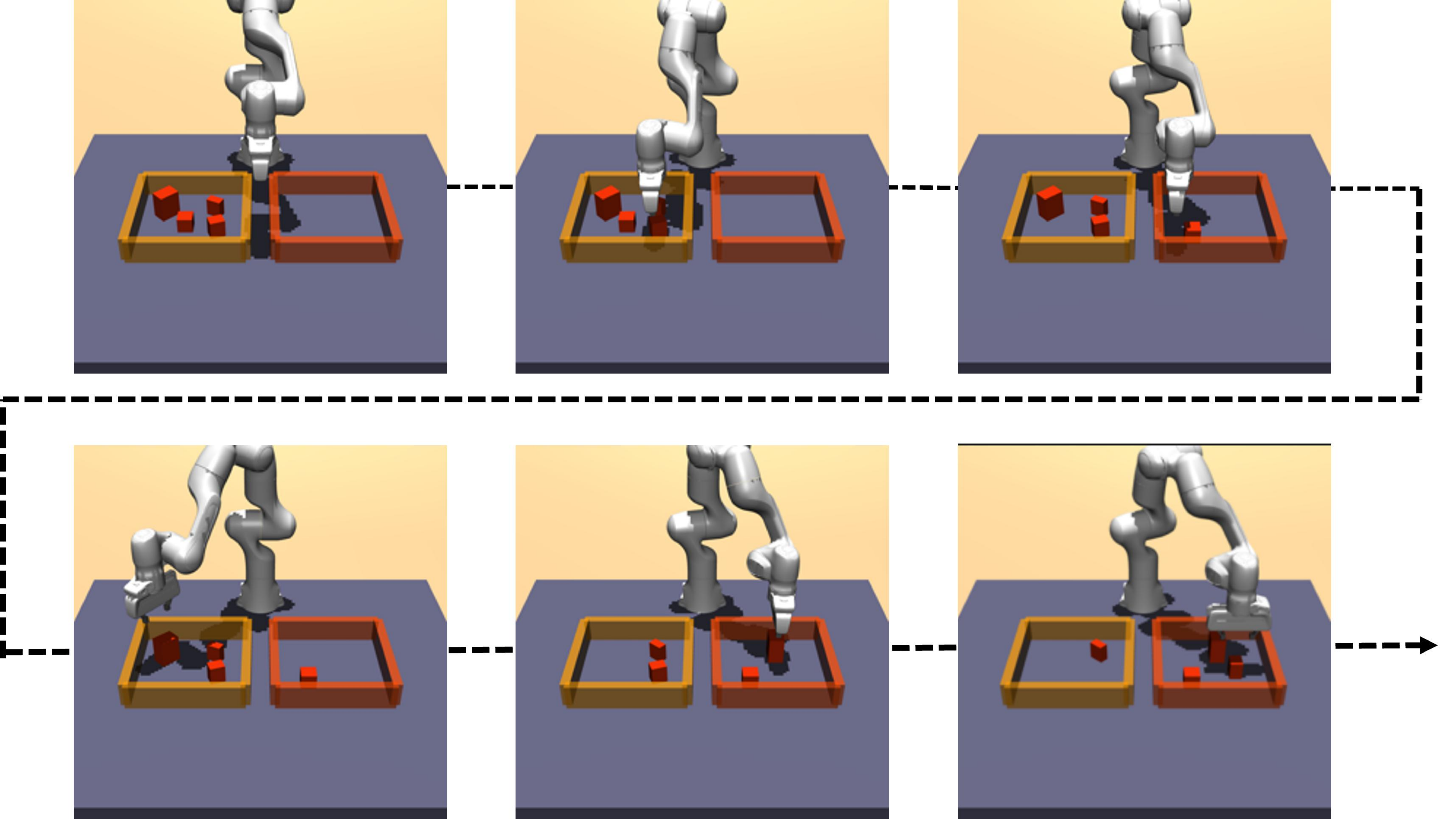}
    \caption{An example of our replacement-task dataset.}
    \label{fig:data}
\end{figure}

\subsubsection{ATC:~Action-Transition Consistency}
Action-transition consistency (ATC) maintains consistency between actions $\act_t$ and transition directions $\dir_t$.
This is inspired by the simple idea that two transition directions should be the same or similar when two corresponding actions are the same or similar. This idea further improves future-prediction performance, as discussed in \cref{sec:cls_exp}.

ATC uses two variances: 1) within action category variance $S_w$ and 2) between action category variance $S_b$, as shown in \cref{fig:atc}; note that $\act_t$ includes an action category, such as grasp, move, and release.
The $S_w$ is used to making transitions in the same action category similar, while $S_b$ is used to have the opposite effect, i.e., making transitions in the different categories dis-similar.

For simplicity, we assume that the same action category is ordered consecutively in an action sequence $\{\act_t\}_{t=1}^T$ and that once a category appears, it will not appear again. Let $C$ be the number of action categories, ${\cal A}_i$ be the temporal index set of the $i$-th action category, and $\mub_i = \frac{1}{|{\cal A}_i|}\sum_{t\in{\cal A}_i}{\dir_t}$, definition of within action category, and between action category variances are as follows:
\begin{align}
    S_w&=\frac{1}{C}\sum_{i=1}^C\frac{1}{|{\cal A}_i|}{\sum_{t\in{\cal A}_i}{\cos(\mub_i, \dir_t)}}=\sum_{i=1}^C{S_w^i} , \\
    S_b&=\frac{1}{C}\sum_{i=1}^{C-1}{\sum_{j=i+1}^C{\cos(\mub_i, \mub_j)}}=\sum_{i=1}^{C-1}{\sum_{j=i+1}^C{S_b^{ij}}},
\end{align}
where $\cos$ is a function to calculate the cosine similarity.
If a sequence of an action category appears multiple times in an overall action sequence, $\mub_i$, $S_w^i$, and $S_b^{ij}$ are calculated with each category sequence. 

ATC is defined as
\begin{align}
    {\rm ATC}: R_{ATC}=S_b-S_w.
    \label{eq:atc}
\end{align}
By minimizing \cref{eq:atc}, the consistency between actions and transition directions is maintained.

\section{Experiments}
\label{sec:exp}
We first demonstrate the effectiveness of our FIRP and TCR through a success-or-failure classification experiment using three public and two self-collected datasets. We then conducted a robotic-manipulation experiment to verify the effectiveness of FIRP in terms of the task-success rate, on our environments. To generate an action plan, we combined FIRP with a TAMP method~\cite{takano2021continuous}.

\subsection{Classification Experiments}
\label{sec:exp_cls}
\begin{table}[t]
    \centering
    \caption{Parameter list tuned by Optuna. $\lambda,\alpha$, and $\beta$ are weights for latent overshooting (\cref{eq:loss}), TTC and ATC (\cref{eq:tcr}), respectively. $\#E_{start}^f$ and $\#E_{stop}^e$ are start epoch of using the future prediction loss and stop epoch of updating ResNet50. }
    \begin{tabular}{rc}
        \toprule
        Learning rate & $10^{-5},10^{-4}$ \\
        Gradient clipping & $10^{-1},10^{0},10^{1},10^{5}$\\
        $\lambda,\alpha,\beta$ & $10^{-3},10^{-2},10^{-1},0$ \\
        $\#E_{start}^f$ & 1, 50, 100, 150 \\
        $\#E_{stop}^e$ & 50, 100, 150, 200 \\
        \bottomrule
    \end{tabular}%
    \label{table:params}
\end{table}

\subsubsection{Experimental Setting} 
\noindent {\bf Data collection:} We used MimicGen~\cite{mimicgen} and self-collected datasets in the classification experiment.

\begin{table*}[tb]
    \centering
    \caption{Balanced accuracies in the classification experiment.}
    \begin{tabular}{lrrrrr}
    \toprule
    ~&  \multicolumn{3}{c}{MimicGen}& \multicolumn{2}{c}{Ours}\\
    \cmidrule(l{.9em}r{.9em}){2-4}\cmidrule(l{.9em}r{.9em}){5-6}
    Task&Three Pieces Assembly& Pick \& Place&Stacking& Replacement &  Stacking\\
    \midrule
    \midrule
    Oracle  & 1.000{$\pm$0.000}&0.981{$\pm$0.016}& 1.000{$\pm$0.000}&0.963{$\pm$0.018}& 0.987{$\pm$0.015}\\
    \midrule 
    A-MLP   & 0.605$\pm$0.042 & 0.817$\pm$0.023 & 0.583$\pm$0.356 &\second{0.655{$\pm$0.019}}&  0.806{$\pm$0.015}\\       
    A-MLP+ResNet& \first{0.938$\pm$0.065} & 0.882$\pm$0.024 & 0.889$\pm$0.060&0.637{$\pm$0.029} &  \second{0.808{$\pm$0.024}}\\
    GRU~\cite{gru} & 0.905$\pm$0.034 & \second{0.926$\pm$0.028}  & \second{0.903$\pm$0.052}&0.561{$\pm$0.059}&  0.755{$\pm$0.041}\\       
     DVD~\cite{dvd}& 0.561$\pm$0.010 &0.685$\pm$0.078 & 0.494$\pm$0.035 & 0.497{$\pm$0.026}& 0.592{$\pm$0.038} \\
    \midrule
    FIRP & \second{0.932$\pm$0.034} & \first{0.952$\pm$0.023} & \first{0.923$\pm$0.028}&\first{0.663{$\pm$0.042}}& \first{0.830{$\pm$0.021}}\\  
    
    \bottomrule
    \end{tabular}%
    \label{tab:acc_results}
\end{table*}

\begin{table}[tb]
    \centering
    \caption{Balanced accuracies on our datasets in the ablation study on temporal consistency regularization (TCR), which consists of TTC and ATC.}
        \begin{tabular}{lrr}
        \toprule
        ~ & Replacement& Stacking  \\
        \midrule
        FIRP  & 0.640{$\pm$  0.047} & 0.788{$\pm$  0.019}\\       
        FIRP + TTC & 0.629{$\pm$  0.028} & \first{0.831{$\pm$  0.018}} \\       
        FIRP + ATC  & \first{0.671{$\pm$  0.033}}  & 0.821{$\pm$  0.023} \\       
        FIRP + TCR  & \second{0.648}{$\pm$  0.046} & \second{0.828}{$\pm$  0.029} \\       
        \bottomrule
        \end{tabular}%
    \label{tab:reg_eff}
\end{table}

\noindent
-~{\bf MimicGen datasets}~\cite{mimicgen}:~We used three datasets: each dataset is a data from three-piece assembly task, pick-place task, and stacking task.
The three-piece assembly task involves inserting two randomly placed pieces into the base piece sequentially, the pick-place task involves placing four randomly placed objects into target places, and the stacking task involves stacking three colored blocks randomly placed on a table.
We used 1000 image sequences and action plans created by an image-based policy trained on the MimicGen system.
The length of each action plan was set to 100, 200, and 80 at 4FPS for three-piece assembly, pick-place, and stacking tasks, respectively.
Please refer to \cite{mimicgen} for the other details of data and tasks, as we refrain from detailing them in this paper due to page limitations.

\noindent
-~{\bf Our datasets}:~
We created two datasets using the MuJoCo simulator~\cite{mujoco}: the first dataset is the data from a replacement task, and the second is the data from a stacking task. The replacement task involves moving three blocks randomly placed in a box into another box, and the stacking task involves stacking three blocks randomly placed on a table. 
\Cref{fig:data} shows an example image sequence of the replacement task.

We generated the action plan introduced by Takano et al.~\cite{takano2021continuous} and collected images and success-or-failure labels automatically by executing the generated plan. The length of each action plan was set to 120 at 4FPS. One action $\act_t$ in the plan consists of a one-hot vector representing an action category and signals to robotic arm joints. There are seven action categories: Move, Grasp, Release, Hold, Move before grasp, Hold before release, and Move after release. We used the Panda arm; thus, there were nine action signals: seven for joint angles and the remaining two signals for grippers. 

The initial position of the blocks was randomly varied, and 360 sequences were collected for the replacement task and 640 sequences for the stacking task.

\noindent {\bf Evaluation protocol:} We randomly selected 80\% of the sequences for training and the rest for testing. We used 20\% of the training data for validation. We calculated the balanced accuracy for test data as an evaluation index. This procedure was repeated three times for the MimicGen datasets and five times for our datasets. We report the average balanced accuracies.

\noindent {\bf Implementation details:} FIRP used ResNet50~\cite{resnet} pretrained on ImageNet~\cite{imagenet} as the image feature $\e_t$ extractor, and gated recurrent unit (GRU)~\cite{gru} as the recurrent neural network in RSSM to predict $\hvec_t$ (\cref{eq:rssm}). We trained FIRP for 300 epochs using Adam optimizer~\cite{adam}. We used Optuna~\cite{optuna} for hyper-parameter tuning. We tuned seven parameters shown in \cref{table:params}.

We selected the top-five models using the validation data and used an average score of their output score for the classification.

\noindent {\bf Baselines:} We compared FIRP with five baselines: Action-MLP (A-MLP), A-MLP+ResNet, GRU, Domain-agnostic Video Discriminator (DVD)~\cite{dvd}, and Oracle. For these baselines other than DVD, we carefully tuned the learning rate and number of epochs.
For DVD, we used the trained model published by the authors of DVD.

\noindent
-~{\bf A-MLP}:~An action plan only. 
A-MLP executes success-or-failure classification with a three-layer MLP by using only an action plan as input to the MLP. 

\noindent
-~{\bf A-MLP+ResNet}:~An action plan and initial image are used without future prediction.
A-MLP+ResNet is an extension of A-MLP that uses an image feature of an initial image. A-MLP+ResNet inputs the image feature extracted by ResNet50 and the action plan to the three-layer MLP. 

\noindent
-~{\bf GRU}:~Future-predictive baseline.
GRU is a method in which the future prediction part of FIRP is replaced with a simple GRU. Thus, GRU is most related to FIRP. 

\noindent
-~{\bf DVD~\cite{dvd}}:~State-of-the-art reward function baseline.
DVD is a method to calculate a task reward while referring to sequences of success trials.
DVD has a high generalization ability, as the DVD model was trained with diverse datasets, including human and robotic videos.
We choose a reference sequence from the training dataset based on the evaluation of the validation dataset, and use actual images for input of the model as in Oracle.

\noindent
-~{\bf Oracle}:~A baseline to show the upper limit of the classification performance as it uses actual image features, not predicted features.
Oracle is a method for executing classification after executing an action plan. That is, the oracle receives an image sequence as an input and uses ResNet as a feature extractor and a two-layer MLP for the classification.

\subsubsection{Results}
\label{sec:cls_res}
{\noindent \bf Overall results:} \cref{tab:acc_results} shows the comparative results of all methods. 
The results support the effectiveness of our method, as our method achieved competitive results compared with the baselines other than Oracle, which is a method that shows the upper limit of the classification performance.
Oracle achieved almost perfect classification results. This suggests that image features are essential for the classification, and the higher the future-prediction accuracy of image features, the higher the classification accuracy.

FIRP’s superiority to A-MLP and A-MLP+ResNet also supports the importance of using predicted image features. FIRP also showed better results than GRU, which is also a future-predictive method. This verifies the advantage of FIRP and its learning algorithm in long-horizon tasks.
FIRP overcame DVD, a state-of-the-art general reward function learned on large-scale datasets.
This supports that FIRP has high adaptability for long-horizon tasks, unlike the simple application of DVD.

Our datasets' classification tasks seem more complex than the MimicGen datasets' tasks due to the relatively low classification performances.
The reason may be that the TAMP planner outputs a plan similar to the successful data, even if the output plan fails a given task, as the TAMP planner used to create our datasets optimizes the entire action plan.
FIRP achieved competitive results in such complex situations, again confirming the effectiveness of our method.

\begin{figure}[tb]
    \centering
    \begin{minipage}[b]{0.45\linewidth}
        \centering
        \includegraphics[width=0.93\linewidth]{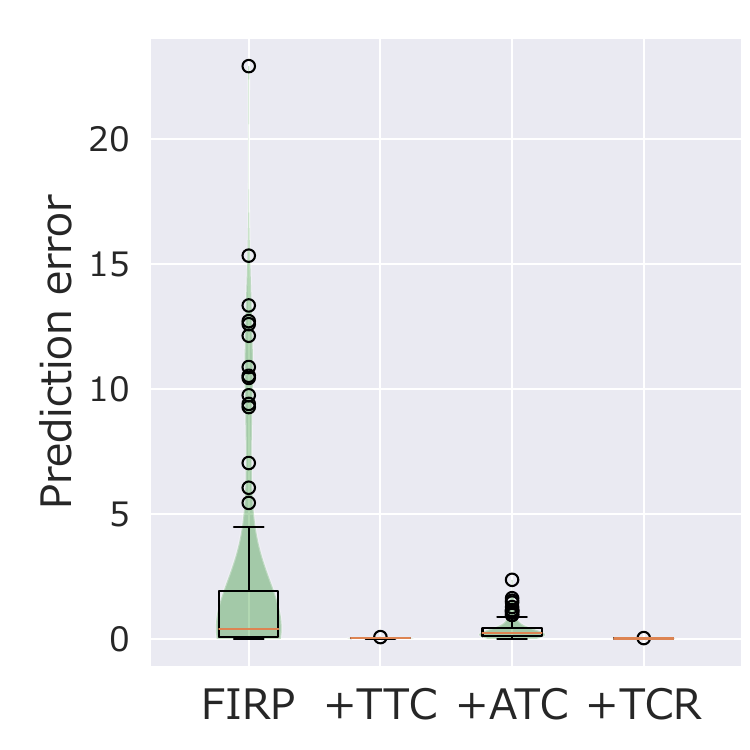}
        \subcaption{Whole figure}
        \label{fig:mse_a}
    \end{minipage}~~
    \begin{minipage}[b]{0.45\linewidth}
        \centering
        \includegraphics[width=0.93\linewidth]{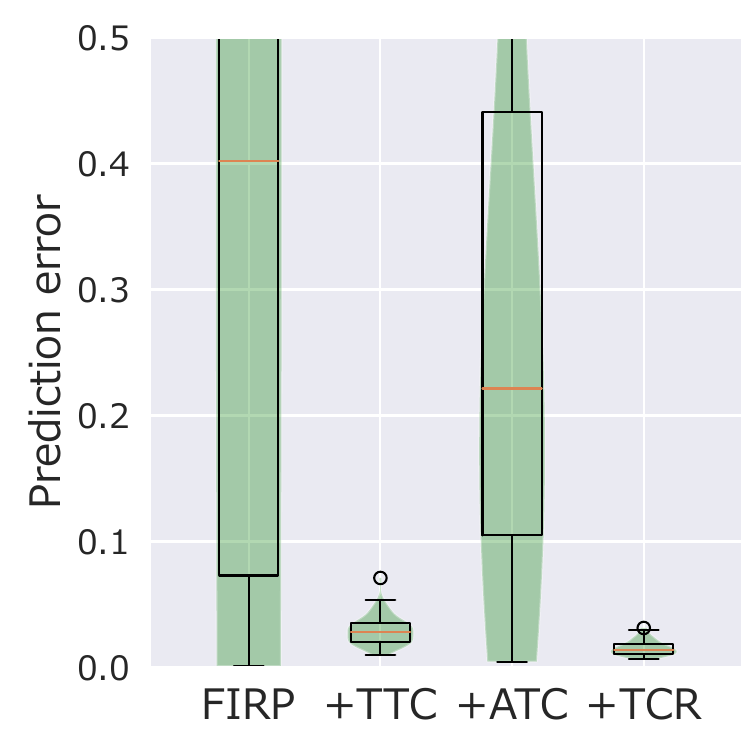}
        \subcaption{Enlarged figure}
        \label{fig:mse_b}
    \end{minipage}
    \caption{Box and violin plots of future prediction errors. We used Euclidean distance as error index. We focused on last time in each sequence, i.e., we calculated Euclidean distance between last image feature $\e_T$ and the corresponding feature ${\hat\e}_T$ predicted from an initial image.}
    \label{fig:mse}
\end{figure}

\begin{figure*}[tb]
    \centering
    \begin{minipage}[b]{0.235\linewidth}
        \centering
        \includegraphics[width=0.85\linewidth]{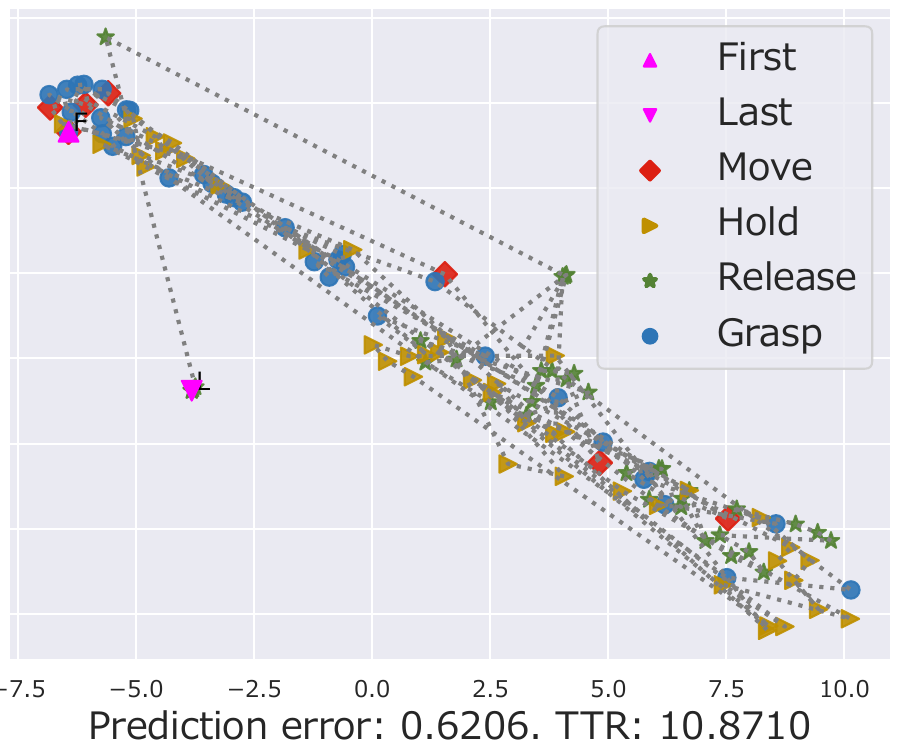}
        \subcaption{FIRP}
        \label{fig:dist_a}
    \end{minipage}
    \begin{minipage}[b]{0.235\linewidth}
        \centering
        \includegraphics[width=0.85\linewidth]{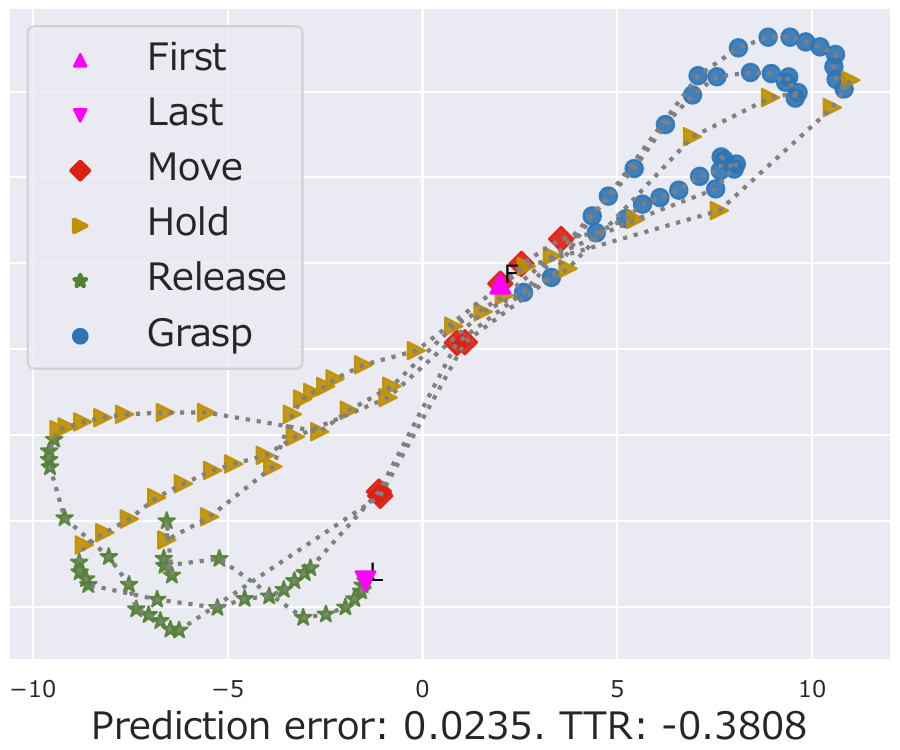}
        \subcaption{FIRP~+~TTC}
        \label{fig:dist_b}
    \end{minipage}
    \begin{minipage}[b]{0.235\linewidth}
        \centering
        \includegraphics[width=0.85\linewidth]{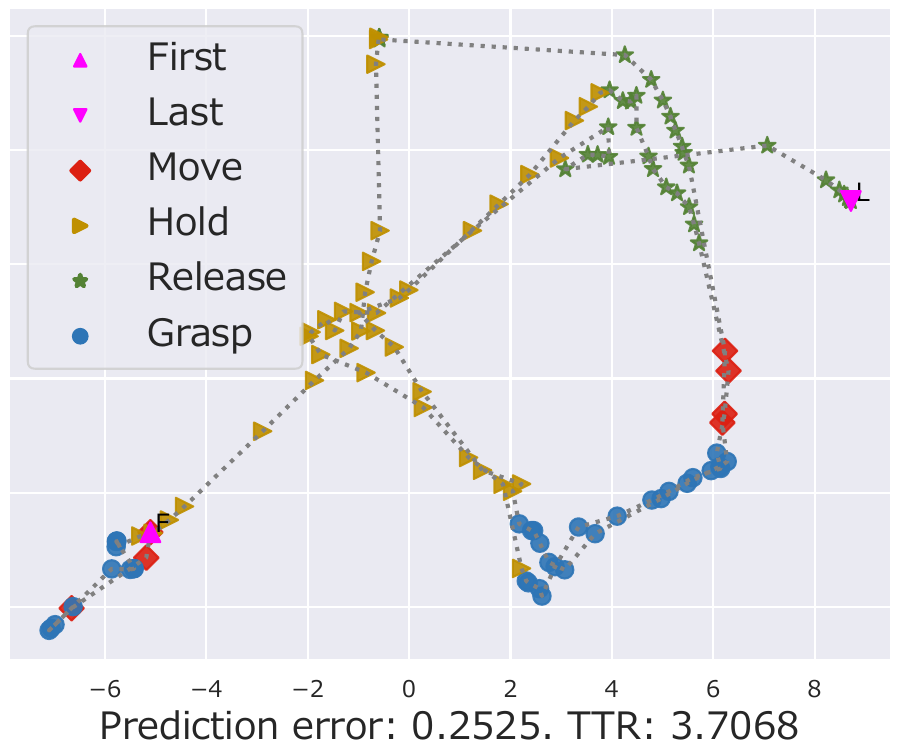}
        \subcaption{FIRP~+~ATC}
        \label{fig:dist_c}
    \end{minipage}
    \begin{minipage}[b]{0.235\linewidth}
        \centering
        \includegraphics[width=0.85\linewidth]{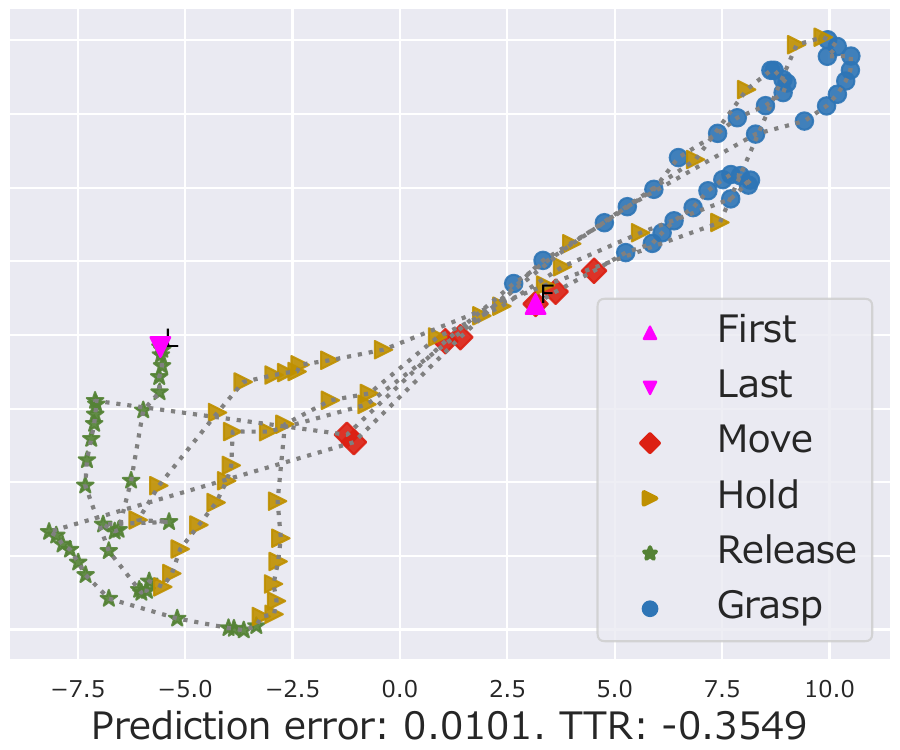}
        \subcaption{FIRP~+~TCR}
        \label{fig:dist_d}
    \end{minipage}
    \caption{Visualization of a temporal transition of image features by t-SNE. Each point (image feature) is colored according to the action category.}
    \label{fig:dist}
\end{figure*}

{\noindent \bf Effectiveness of Regularization:}
\label{sec:cls_exp}
~\Cref{tab:reg_eff} shows the classification results in the ablation study on our regularization term, TCR, which consists of two consistencies: 1) temporal transition consistency (TTC) and 2) action-transition consistency (ATC).
Overall, the regularization improves classification accuracies.
Although the best method depends on tasks, FIRP+TCR achieved high classification results stably.
This suggests that our key ideas, maintaining consistency in temporal transition and consistency between actions and transition directions, are efficient for improving classification and future-prediction performance.
 
\Cref{fig:mse} shows future prediction errors.
TTC and ATC improve future prediction performance, and the combination of TTC and ATC, i.e., TCR further improves prediction performance.
This improvement may enhance the stability of classification performance.

\Cref{fig:dist} shows a transition of an image feature sequence using t-SNE~\cite{tsne}.
\Cref{fig:dist_b,fig:dist_c,fig:dist_d} show that image features move smoothly and are well localized according to an action category, compared with \cref{fig:dist_a}.
These figures prove that TCR realizes the following key ideas: feature transitions can be decomposed into elements common to the temporal direction and action categories, making a feature distribution easy-to-predictable.

\subsection{Robotic Manipulation Experiment}
\label{sec:exp_robo}

\subsubsection{Experimental Setting}
\label{sec:robo_set}
In this experiment, we evaluated the combination of FIRP with the optimization-based planning method~\cite{takano2021continuous} in terms of success rate in the replacement and stacking tasks.

{\noindent \bf Planning with TAMP and FIRP:} We used a TAMP method~\cite{takano2021continuous} as the baseline. 
As a simple combination method of FIRP with the TAMP~\cite{takano2021continuous}, we carried out re-planning by the TAMP when our method identifies that the execution of an action-plan candidate would fail a given task.
Before executing re-planning, we added a condition of changing the moving order of blocks to generate a different action plan.

{\noindent \bf Implementation details:} The iteration of (re-)planning and success-or-failure classification was repeated until FIRP classified a candidate action plan as successful, while the maximum number of iterations was set to six.
We executed an action plan when the maximum number of iterations was reached, even if FIRP identified it as failure.
We conducted 50 task trials and calculated the task success rate using the results of the trials.
We use the proposed model with the best validation accuracy, evaluated in the same protocol as the previous experiment.

\subsubsection{Results}
\label{sec:robo_res}
\begin{table}[t]
    \centering
    \caption{Task success rates in the robotic manipulation experiments.  The success rates were calculated on the results of 50 trials.}
    \begin{tabular}{lrr}
        \toprule
        & Replacement  & Stacking \\
        \midrule
        TAMP~\cite{takano2021continuous}& 56 & 34  \\
        \hdashline
        \:\:\:\:\:\:+~FIRP & 60 & 76  \\
        \:\:\:\:\:\:+~FIRP+TTC & 56 & 68 \\
        \:\:\:\:\:\:+~FIRP+ATC & 62 & 74 \\
        \:\:\:\:\:\:+~FIRP+TCR & 62 & 86 \\
        \bottomrule
    \end{tabular}%
    \label{table:plan_results}
\end{table}

\begin{table}[t]
    \centering
    \caption{Task success rates in the robotic manipulation experiments. The success rates were calculated only on data FIRP identified as successful.}
    \begin{tabular}{lrr}
        \toprule
         & Replacement & Stacking  \\
        \midrule
        TAMP~\cite{takano2021continuous}& 56  & 34 \\
        \hdashline
        \:\:\:\:\:\:+~FIRP & 64  & 83 \\
        \:\:\:\:\:\:+~FIRP+TTC & 67 & 92 \\
        \:\:\:\:\:\:+~FIRP+ATC & 65 & 77  \\
        \:\:\:\:\:\:+~FIRP+TCR & 62 & 86 \\
        \bottomrule
    \end{tabular}%
    \label{table:plan_results_success}
\end{table}

\Cref{table:plan_results} shows the comparative results.
We can see that re-planning with our methods significantly improves task success rates.
The low success rate of TAMP suggests that conditions required by TAMP might be overlooked in the manual design process. 
The improvement by introducing FIRP suggests that FIRP automatically obtained overlooked conditions and could give feedback to TAMP.

The results in \cref{table:plan_results} are calculated from all trials, even if FIRP finally identified an action plan will fail.
\Cref{table:plan_results_success} shows success rates calculated only on trials FIRP identified as successful.
We can see that success rates are further improved.
This result verifies the effectiveness of FIRP's success-or-failure classification performance.

Those results also suggest that our method has two possible use cases depending on applications.
The first use case is to improve feasibility of an action plan and execute the improved plan for applications requiring enormous manipulations.
The second use case is to inform operators of possible failures when FIRP identifies a plan will fail after improving the feasibility of an action plan, for applications with high responsibility.

\section{Conclusion}
\label{sec:conclusion}
We proposed a future-predictive success-or-failure classification method for long-horizon tasks, which is an alternative to the manual design of conditions required by optimization-based planning methods. 
The key part of our method is long-horizon prediction.
We used a recurrent state space model and proposed a regularization called transition consistency regularization (TCR). This term is designed to maintain temporal transition and action-transition consistencies; these consistencies were defined with temporal transition directions. The results from classification and robotic-manipulation experiments indicate that our method achieved high classification performance and improved the success rate of long-horizon tasks. We believe that this paper will provide a practical direction of automating long-horizon tasks and a novel research direction of applying a machine-learning approach to robotic tasks.

\bibliographystyle{IEEEtran}
\bibliography{ref}

\end{document}